%% file: main.tex
\documentclass{article}

\usepackage{bmpsize}
\usepackage[dvipdfmx]{graphicx}

\input{math_commands.tex}

\usepackage{url}
\usepackage{subcaption}
\usepackage{multirow}
\usepackage{adjustbox}
\usepackage{natbib}

\usepackage{microtype}
\usepackage{graphicx}
\usepackage{booktabs} 

\usepackage{hyperref}

\usepackage[accepted]{icml2025}

\usepackage{amsmath}
\usepackage{amssymb}
\usepackage{mathtools}
\usepackage{amsthm}

\usepackage[capitalize,noabbrev]{cleveref}

\theoremstyle{plain}

\theoremstyle{definition}

\theoremstyle{remark}

\usepackage[disable,textsize=tiny]{todonotes}

\icmltitlerunning{ComMer: a Framework for Compressing and Merging User Data for Personalization}

\begin{document}

\twocolumn[
\icmltitle{ComMer: a Framework for Compressing and Merging \\
           User Data for Personalization}




\begin{icmlauthorlist}
\icmlauthor{Yoel Zeldes}{google}
\icmlauthor{Amir Zait}{google}
\icmlauthor{Ilia Labzovsky}{google}
\icmlauthor{Danny Karmon}{google}
\icmlauthor{Efrat Farkash}{google}
\end{icmlauthorlist}

\icmlaffiliation{google}{Google DeepMind}

\icmlcorrespondingauthor{Yoel Zeldes}{yoelz@google.com}
\icmlcorrespondingauthor{Amir Zait}{amirzait@google.com}
\icmlcorrespondingauthor{Ilia Labzovsky}{ilabz@google.com}
\icmlcorrespondingauthor{Danny Karmon}{dannykarmon@google.com}
\icmlcorrespondingauthor{Efrat Farkash}{efratf@google.com}

\icmlkeywords{ICML, LLM, compression, merging, personalization, LaMP}

\vskip 0.3in
]



\printAffiliationsAndNotice{}  

\begin{abstract}
Large Language Models (LLMs) excel at a wide range of tasks, but adapting them to new data, particularly for personalized applications, poses significant challenges due to resource and computational constraints. Existing methods either rely on exposing fresh data to the model through the prompt, which is limited by context size and computationally expensive at inference time, or fine-tuning, which incurs substantial training and update costs. In this paper, we introduce ComMer - Compress and Merge - a novel framework that efficiently personalizes LLMs by compressing users' documents into compact representations, which are then merged and fed into a frozen LLM. We evaluate ComMer on two types of personalization tasks - personalized skill learning, using the tweet paraphrasing dataset and the personalized news headline generation dataset from the LaMP benchmark, and knowledge-intensive, using the PerLTQA dataset. Our experiments demonstrate that in constrained inference budget scenarios ComMer achieves superior quality in skill learning tasks, while highlighting limitations in knowledge-intensive settings due to the loss of detailed information. These results offer insights into trade-offs and potential optimizations in multi-document compression for personalization.
\end{abstract}

\input{01_intro}
\input{02_related}
\input{03_method}
\input{04_experimental_setup}

\input{05_results}
\input{06_discussion}
\input{07_impact}

\bibliography{bibliography}
\bibliographystyle{icml2025}

\newpage
\appendix
\clearpage
\input{08_appendix}

\end{document}

%% file: math_commands.tex
\usepackage{amsmath,amsfonts,bm}

\def\eqref#1{equation~\ref{#1}}

\def\1{\bm{1}}

\DeclareMathAlphabet{\mathsfit}{\encodingdefault}{\sfdefault}{m}{sl}
\SetMathAlphabet{\mathsfit}{bold}{\encodingdefault}{\sfdefault}{bx}{n}



%% file: 01_intro.tex
\section{Introduction}

LLMs have demonstrated remarkable capabilities across a wide range of natural language processing tasks. In many tasks, a simple prompt suffices to guide the model toward the desired behavior, while leveraging any required knowledge acquired during its pretraining \cite{NEURIPS2020_1457c0d6}. However, in some cases, such as personalizing LLMs to users' data, new data becomes available only after the model’s initial training. Adapting LLMs to accommodate this new data introduces additional resource requirements and computational overhead, which is significant when scaling to a large number of users.

\begin{figure*}
\vskip 0.2in
\begin{center}
\includegraphics[width=0.7\linewidth]{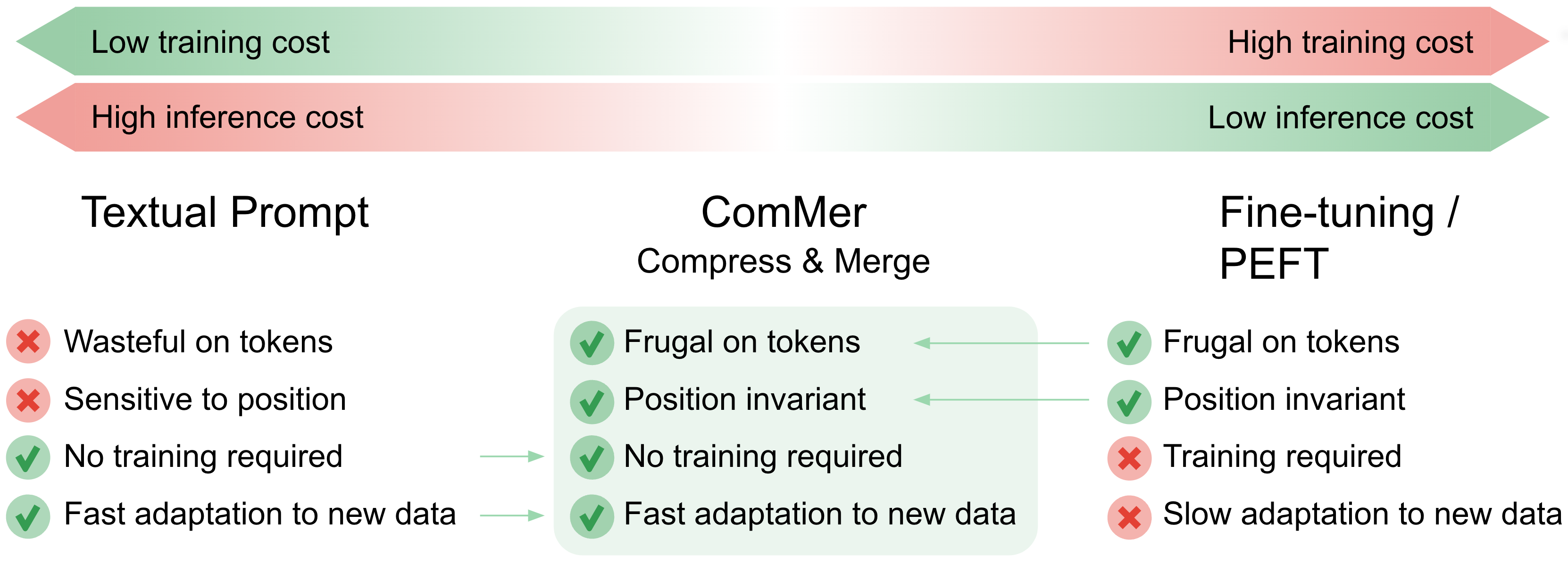}
\caption{Approaches for adapting LLMs to new data include integrating it through the prompt or modifying the model by updating its existing weights or introducing new trainable weights. Both methods have advantages and drawbacks, while our proposed method, ComMer, combines the benefits of both approaches.}
\label{fig:motivation}
\end{center}
\vskip -0.2in
\end{figure*}

The two prevalent approaches for adapting LLMs to new data, as illustrated in Figure \ref{fig:motivation}, are (1) prompt engineering, where new data is incorporated into the model through the prompt \cite{salemi2024comparingretrievalaugmentationparameterefficientfinetuning}, and (2) full or parameter-efficient fine-tuning (PEFT) \cite{wang2024parameterefficientfinetuninglargemodels}. The first approach faces limitations due to the model’s supported context window. While recent advancements allow LLMs to process millions of tokens \cite{geminiteam2024gemini15unlockingmultimodal}, this becomes problematic as the number and length of documents increase — particularly in data-rich modalities like video and audio. Even when the context window is sufficient, LLMs often exhibit positional bias, processing documents differently depending on their placement within the prompt, which can affect output quality \cite{liu-etal-2024-lost}. Additionally, regardless of quality, both latency and computational costs scale with the length of the prompt \cite{pmlr-v201-duman-keles23a}, making this approach less feasible for large-scale applications.

In the second approach, the LLM is adapted to new data by modifying its weights — or a subset of them — to incorporate the information from the new data. While this addresses the limitations of prompt engineering, it introduces new challenges. Specifically, training a separate model for each user leads to substantial computational costs. Furthermore, updating personalized models as new user data becomes available is a resource-intensive and complex process, often necessitating offline updates that compromise the model's freshness. Finally, training personalized models for users with limited data can result in low-quality models \cite{zhuang2024hydramodelfactorizationframework}.

In this paper, we propose ComMer - Compress and Merge - a novel approach designed to address these challenges by learning to compress user data into a compact representation that efficiently adapts the LLM. The cost of training ComMer is amortized across all users, eliminating the need for individualized training processes. This approach offers several advantages, including reduced computational costs, streamlined and cost-effective adaptation to new data, and improved quality for users with limited data.

We evaluate ComMer on two types of personalization tasks: personalized skill learning, using the tweet paraphrasing dataset and the personalized news headline generation dataset from the LaMP benchmark \cite{salemi-etal-2024-lamp}, and knowledge-intensive question answering, using the PerLTQA dataset \cite{du-etal-2024-perltqa}. Our experiments demonstrate that ComMer strikes an effective balance between quality and computational cost in personalized skill learning. It efficiently processes multiple documents under a constrained token budget, with quality improving as the number of documents grows. However, in knowledge-intensive tasks, ComMer struggles to represent all the information from the documents, resulting in quality degradation as the document count increases.
Additionally, we examine the effects of pretraining, generalization to out-of-distribution data, and alternative merging strategies, offering insights into the trade-offs and potential optimizations for ComMer's architecture.

%% file: 02_related.tex
\section{Related Work}

\subsection{Personalization}
Personalizing LLMs spans a spectrum from treating user data as a knowledge base to personalized skill learning. \cite{du-etal-2024-perltqa} introduced PerLTQA, focusing on knowledge base personalization by combining synthetic semantic and episodic memories for Q\&A tasks. In contrast, \cite{salemi-etal-2024-lamp} explored personalized skill learning, and introduced the LaMP benchmark consisting of seven personalized tasks. Different methods have been devised to tackle the personalization challenge: \cite{salemi2024comparingretrievalaugmentationparameterefficientfinetuning} evaluated Retrieval-Augmented Generation (RAG) and PEFT on the LaMP benchmark; \cite{10.1145/3626772.3657783} optimized RAG pipelines for retrieval; and \cite{zhuang2024hydramodelfactorizationframework} proposed HYDRA, a model factorization framework with user-specific rerankers and adapters. All these methods incur high inference or training costs.

Concurrent to our work, \cite{liu2024llmspersonaplug} introduced PPlug, which compresses users' documents into pluggable prompts using a frozen encoder. Their approach relies on a query-dependent weighted average of document embeddings, which adds computational and storage overhead since it cannot be precomputed. Furthermore, PPlug's frozen encoder and small embedding dimension limit its capacity, causing embedding interference. In contrast, ComMer uses trainable embeddings with mean pooling, enabling it to co-embed multiple documents in the same latent space while mitigating interference.

\subsection{Efficiency and Compression}
Despite advances in context window sizes \cite{geminiteam2024gemini15unlockingmultimodal}, processing long contexts remains resource-intensive. Compression methods include latent representations \cite{NEURIPS2023_3d77c6dc, chevalier-etal-2023-adapting, ge2024incontext} and natural language compressions \cite{zhou2023efficientpromptingdynamicincontext, jiang-etal-2023-llmlingua, jiang-etal-2024-longllmlingua}. Our approach extends latent representation methods to the multi-document setting by merging document representations, yielding further compression gains.

\subsection{Model Merging}
Merging models enhances their collective performance across tasks, with methods ranging from simple weight averaging \cite{pmlr-v162-wortsman22a, li2022branchtrainmergeembarrassinglyparalleltraining} to sophisticated techniques like spherical linear interpolation \cite{goddard-etal-2024-arcees} and sparse-aware merging \cite{pmlr-v235-yu24p}. Some works focus on merging entire models, while others address merging PEFT models \cite{chitale2023taskarithmeticloracontinual, wu2024mixture, lin2024inducinggeneralizationlanguagestasks, pmlr-v235-ostapenko24a, 10.1007/978-3-031-73232-4_24}. Unlike these approaches, ComMer merges document representations rather than weights, enabling dynamic input adaptation and overcoming the limitation of any available model inventories.

%% file: 03_method.tex
\section{Method}

\subsection{Architecture}

\begin{figure*}
\vskip 0.2in
\begin{center}
\includegraphics[width=1\linewidth]{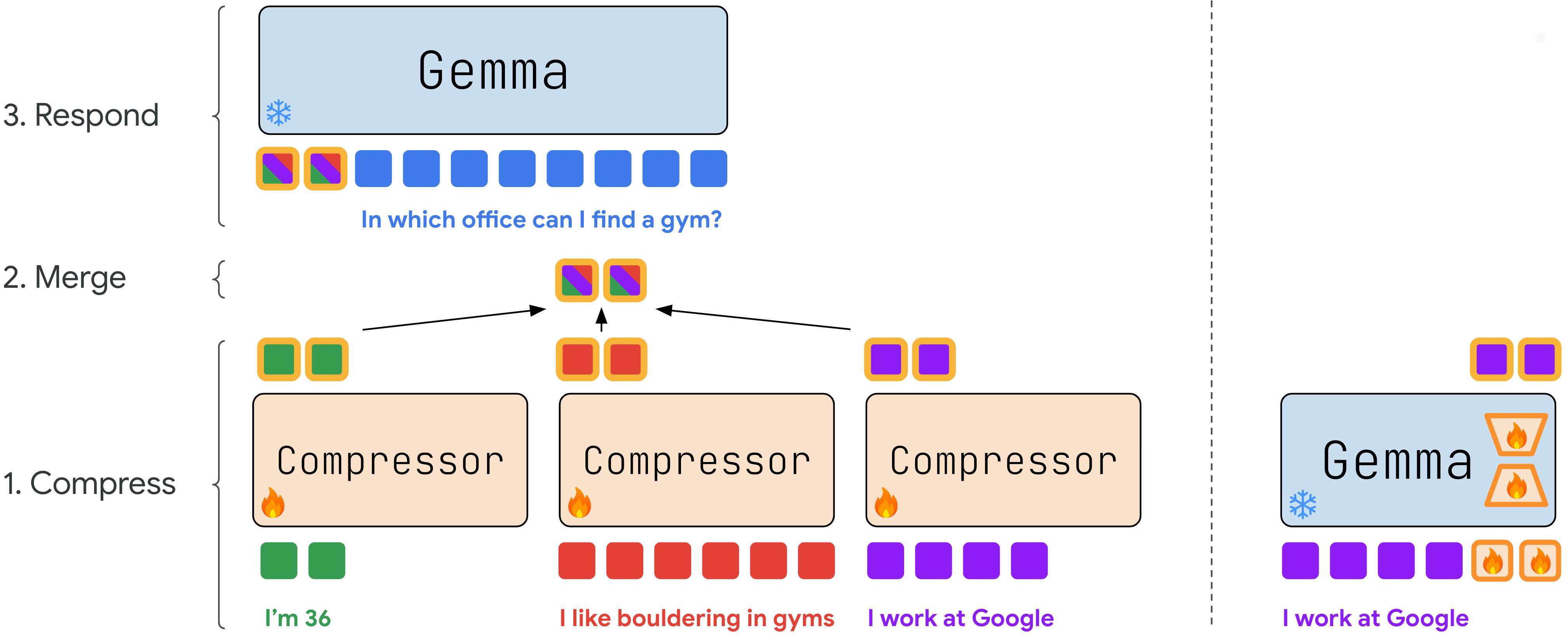}
\caption{\textbf{Left}: ComMer architecture. Each document is independently compressed into a fixed-size representation by a trainable compressor. These compressions are then merged using mean pool. Finally, the aggregated compression is plugged into a frozen LLM.
\textbf{Right}: The compressor architecture. The input is appended with trainable compression embeddings, and processed by a frozen LLM, which is adapted using a trainable LoRA. The compressor's output consists of the final layer’s representations of the compression embeddings.}
\label{fig:architecture}
\end{center}
\vskip -0.2in
\end{figure*}

Figure \ref{fig:architecture} illustrates our proposed architecture. Given a collection of user documents, the process follows three steps:
\begin{enumerate}
    \item \textbf{Compression}: Each document is independently compressed into a soft prompt. Similarly to \cite{ge2024incontext}, the compression is implemented using a frozen LLM with trainable compression embeddings and LoRA weights \cite{hu2022lora}. Specifically:
    \begin{enumerate}
        \item Trainable compression embeddings are appended to the input document.
        
        \item The document, along with the compression embeddings, is passed through a frozen LLM. Leveraging the language understanding of the frozen LLM, which is a required capability of a good document compressor, can reduce training costs. A trainable LoRA adapter is added to adjust the LLM for compression purposes.
        
        \item The last layer's representations of the compression embeddings are extracted as the compressor's output.
    \end{enumerate}
    
    \item \textbf{Merging}: The compressed representations of all documents are aggregated into a single soft prompt using mean pooling. This approach offers several advantages:
    \begin{enumerate}
        \item The shape of the resulting representation is independent of the number of documents, addressing limitations related to the context window and computational cost.
        \item The process is order-agnostic, eliminating the bias related to the position of documents.
        \item The aggregated soft prompt can be precomputed and updated efficiently as new documents are added.
    \end{enumerate}
    
    \item \textbf{Response Generation}: The aggregated soft prompt is plugged into a frozen LLM to generate the desired output.
\end{enumerate}

\subsection{Training}

We assume a dataset of the form $\bigl\{ \langle \{ \text{doc}^i_j \}_{j=1..N_i}, x_i, y_i \rangle \bigl\}_{i=1..N}$ where $N$ is the number of examples, $x_i$ is an instruction, and $y_i$ is the expected response based on the personalized information found in the collection of $N_i$ documents $\{ \text{doc}^i_j \}_{i=1..N_i}$. Each example is processed by our proposed architecture by independently compressing the documents and then passing $x_i$ and $y_i$ as inputs to the frozen LLM, along with the aggregated compression. We employ cross-entropy loss over the tokens of $y_i$, propagating gradients only to the trainable compression embeddings and LoRA weights, while keeping the backbone LLM's weights unchanged.

%% file: 04_experimental_setup.tex
\section{Experimental Setup}

\subsection{Data}
ComMer represents a collection of documents using mean pooling, which makes it an efficient method for handling personalization tasks where user data evolves over time. When a new document is added, only that document needs to be processed, after which it is incorporated into the aggregated compression of the existing document set (with normalization applied to maintain the mean pool's structure). This eliminates the need to reprocess older documents, leading to significant cost savings.

We evaluate ComMer on three such tasks, with examples from each shown in Appendix \ref{section:dataset_examples}.

\subsubsection{Personalized skill learning}
In these tasks, the user's documents provide a signal that is essential for personalizing task performance. To succeed, ComMer must learn to extract the personalization signal from a given set of documents. We chose to experiment with two tasks from the LaMP benchmark: (1) personalized tweet paraphrasing, where the objective is to rephrase a tweet to align with the user’s style, as inferred from their past tweets; and (2) personalized news headline generation, where the objective is to generate a headline for a given article that reflects the author's style, as inferred from their past article-headline pairs. We follow the user-based split described in LaMP \cite{salemi-etal-2024-lamp}, where the test set includes users who were not seen during training.

\subsubsection{Knowledge Intensive}
In these tasks, the documents serve as a personalized knowledge base for each user. To perform well, ComMer must learn to retain the information from these documents, sometimes even verbatim. We chose to experiment with a dataset that is derived from PerLTQA \cite{du-etal-2024-perltqa}, a QA dataset based on synthetic personal memories. For each synthetic persona, their memories, and associated questions and answers, we construct examples by randomly selecting a subset of memories, along with a randomly chosen question and answer. 27 personas are allocated to the training set and 5 to the test set. For simplicity, we overload the name PerLTQA to refer to this processed version of the dataset.

\subsubsection{Pretraining Data}
    Additionally, following \cite{ge2024incontext}, we pretrain ComMer using an unsupervised task. We extend the auto-encoding task to handle the multi-document setup. In the single-document setup, the model is tasked to reconstruct the text from a compressed document. For the multi-document case, we prefix the $i$-th document with "Document $i$:", and use “What does document $n$ contain?” as a query, where $n$ is randomly selected. We utilize documents from the FineWeb dataset \cite{penedo2024finewebdatasetsdecantingweb}, truncating each one to 150 characters. Multiple documents are then grouped to form a single example. We construct a training set containing one million examples. Unless stated otherwise, all models in this research underwent a pretraining phase of one epoch.

\subsection{Models}
We conduct experiments using Gemma-2b \cite{gemmateam2024gemmaopenmodelsbased} as the backbone for both the generator and the compressor, optimizing resource efficiency by sharing weights across the modules. The input to the generator is:

$\text{concat}\bigg(\text{merge}\Big(\big\{\text{compress}(\text{doc}_i)\big\}_{i=1 \dots n}\Big), \, \text{embed}(x)\bigg)$

where $embed$ is the output of the backbone's embedding layer. We use mean pool as the merge operation and experiment with varying the number of compression embeddings. As a baseline, we use prompt-tuning \cite{lester-etal-2021-power}:

$\text{concat}\Big(\text{soft\_prompt}, \, \text{embed}\big(\text{concat}(\text{doc}_1, \dots, \text{doc}_n, \, x)\big)\Big)$

where $soft\_prompt$ represents a list of trainable embeddings. The number of embeddings in the soft prompt is analogous to the number of compression embeddings in ComMer.

Implementation details can be found at \ref{section:implementation_details}.

\subsection{Metrics}

The LaMP benchmark relies on ROUGE-based metrics \cite{lin-2004-rouge} for both the tweet paraphrasing and personalized news headline generation tasks. Our experiments reveal that these metrics exhibit significant noise, which complicates drawing coherent conclusions. Consequently, we incorporate perplexity alongside ROUGE-L in our analysis. Our conclusions are supported by the perplexity results, as well as the ROUGE-L results, albeit with greater variability.
Similarly, for the PerLTQA benchmark, we employ both ROUGE-L and perplexity metrics.

%% file: 05_results.tex
\section{Results}

\subsection{Personalized skill learning}

\begin{figure*}
\vskip 0.2in
\begin{center}
\includegraphics[width=1\linewidth]{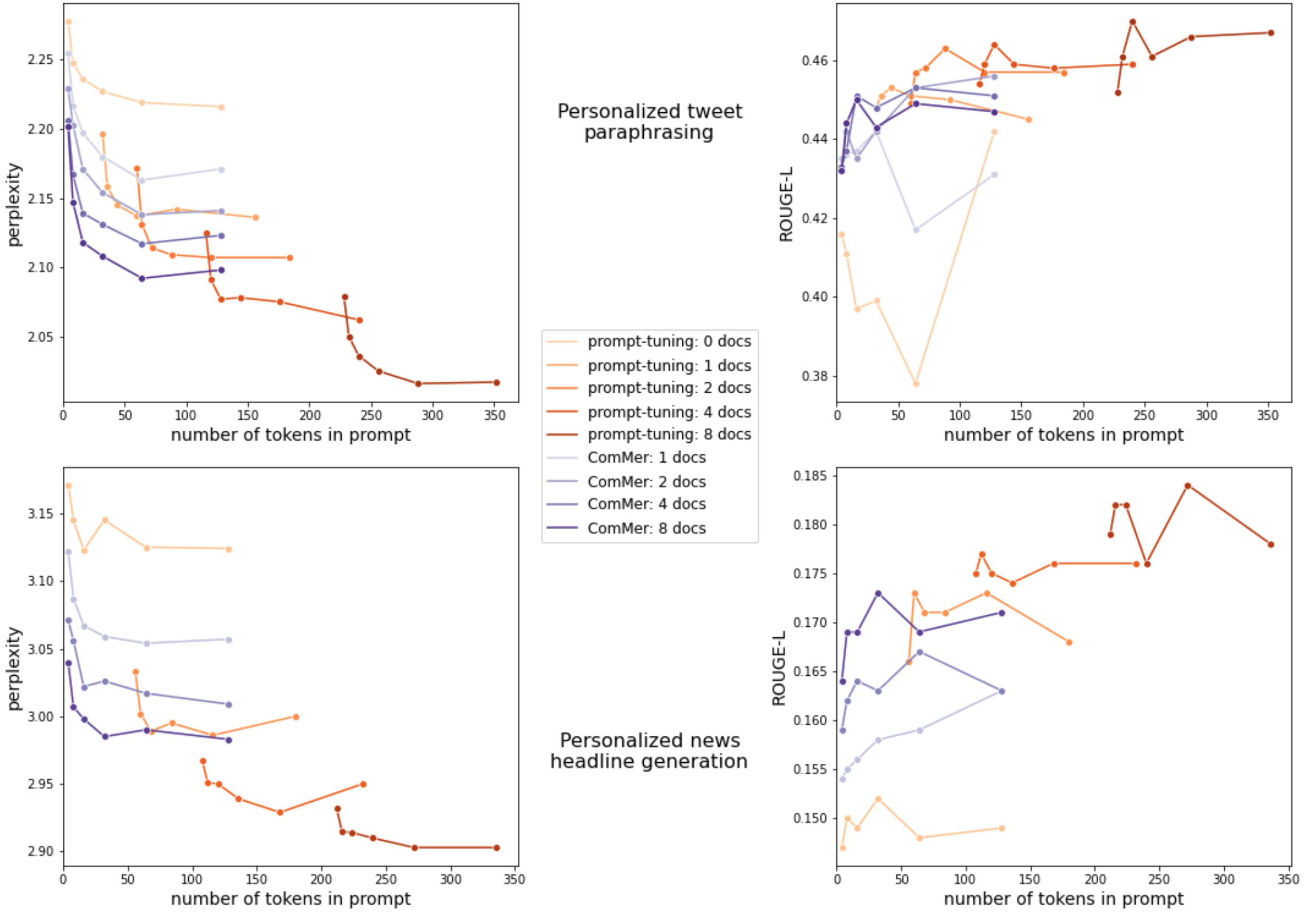}
\caption{The trade-off between cost (number of tokens in the prompt) and quality (perplexity on the left, and ROUGE-L on the right), demonstrated using two personlized skill learning tasks: personalized tweet paraphrasing (top) and personalized news headline generation (bottom). Each curve represents models trained with different numbers of embeddings: 4, 8, 16, 32, 64, and 128, ordered from left to right. In the small token budget regime, ComMer achieves higher quality results with fewer resources than prompt-tuning, highlighting its ability to efficiently extract personalization signals from multiple documents.}
\label{fig:lamp_commer_vs_pt}
\end{center}
\vskip -0.2in
\end{figure*}

\begin{figure*}
\vskip 0.2in
\begin{center}
\includegraphics[width=1\linewidth]{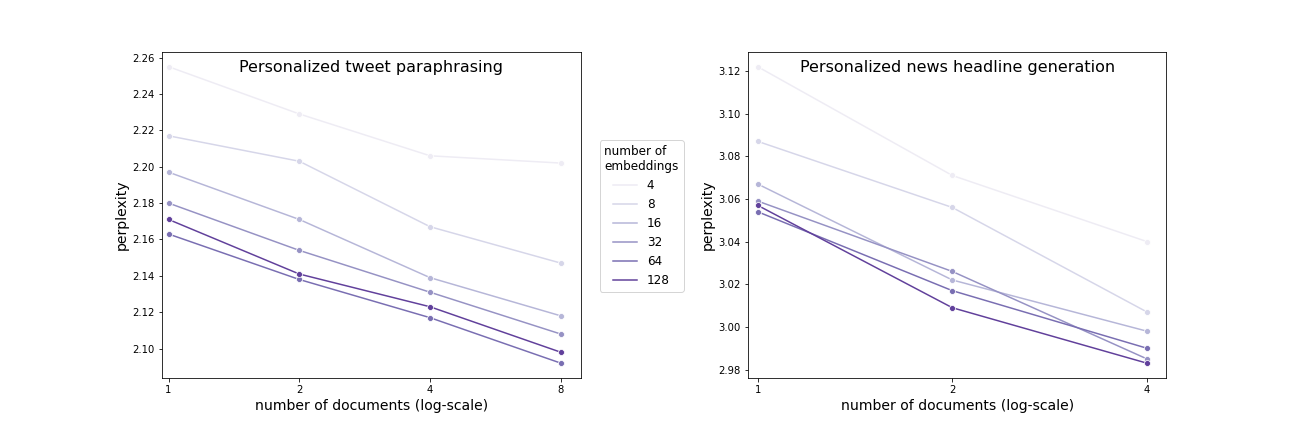}
\caption{Perplexity of ComMer as a function of the number of documents follows a power-law relation in both the personalized tweet paraphrasing (left) and the personalized news headline generation (right) tasks. This pattern holds across all numbers of compression embeddings used by ComMer. It suggests that increasing the number of documents will further enhance ComMer's quality.}
\label{fig:power_law_perplexity}
\end{center}
\vskip -0.2in
\end{figure*}

We trained several ComMer and prompt-tuning models, varying the number of documents (0, 1, 2, 4, 8 for the tweet paraphrasing task, and 0, 1, 2, 4 for the personalized news headline generation task) and the number of embeddings (4, 8, 16, 32, 64, 128). Figure \ref{fig:lamp_commer_vs_pt} illustrates the trade-off between prompt length, which is associated with costs, and model quality. We note that the number of embeddings influences the prompt length, as the embeddings are provided as direct inputs to the frozen LLM.

It is evident that for both prompt-tuning and ComMer, the number of documents and the number of embeddings directly correlate with model quality. For prompt-tuning, they anti-correlate with costs, a relationship that does not apply to ComMer when inspecting the number of documents. Notably, when inspecting the perplexity results of the personalized tweet paraphrasing task, within a budget of up to 128 tokens\footnote{Small token budgets are common in high-throughput applications that generate batches of completions, where GPU memory is scarce.}, ComMer achieves higher quality results while using less resources compared to prompt-tuning, highlighting its ability to extract useful personalization signals from multiple documents within a limited token budget. When inspecting ROUGE-L or the personalized news headline generation task, the token cut-off differs, but still highlights ComMer's effectiveness. We anticipate that more documents can further enhance results, as suggested by the power-law in Figure \ref{fig:power_law_perplexity}. Unfortunately, the datasets we use contain at most eight documents for most users. We leave exploration of other personalized skill learning datasets with a larger document set per user for future work.

\subsection{Knowledge intensive}

\begin{figure*}[t]
\vskip 0.2in
\begin{center}
\includegraphics[width=1\linewidth]{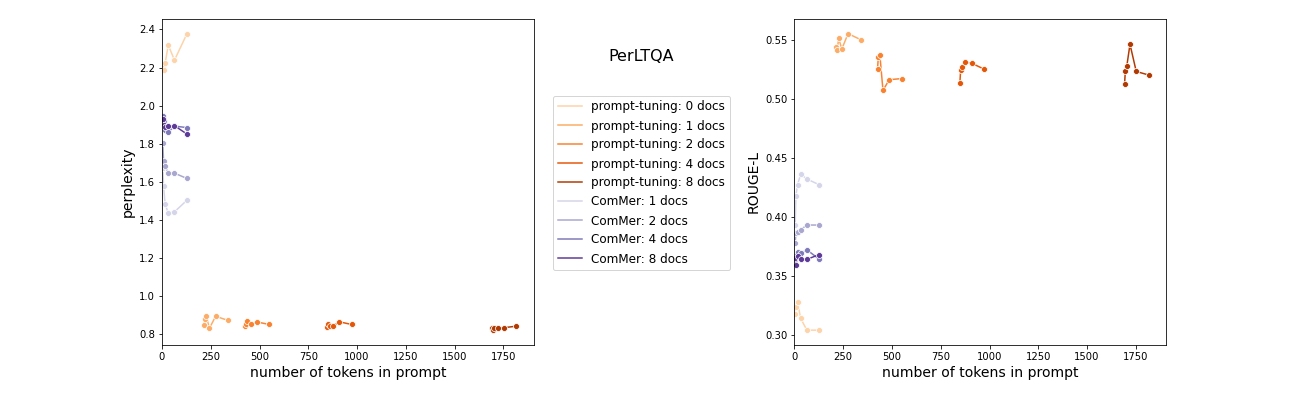}
\caption{The trade-off between cost (number of tokens in the prompt) and quality (perplexity on the left, and ROUGE-L on the right), demonstrated using PerLTQA. Each curve represents models trained with different numbers of embeddings: 4, 8, 16, 32, 64, and 128, ordered from left to right. Compressing multiple documents degrades quality, indicating that ComMer may not be well-suited for the knowledge-intensive nature of PerLTQA.}
\label{fig:perltqa_commer_vs_pt}
\end{center}
\vskip -0.2in
\end{figure*}

In knowledge-intensive tasks, the situation differs significantly, as shown in Figure \ref{fig:perltqa_commer_vs_pt}. In these tasks, having only a single document is the optimal scenario, as additional documents introduce non-relevant information. In typical RAG applications, several documents are retrieved - much like our experimental setup - with the expectation that the LLM can interpret them effectively. As observed, prompt-tuning is robust to irrelevant information, whereas ComMer is not. Unlike in the personalized skill learning setting, the inclusion of more documents leads to worse perplexity and ROUGE-L results for ComMer.

\subsection{Pretraining}

\begin{table*}[t]
    \centering
    \caption{The impact of pretraining on downstream task performance, demonstrated using the personalized tweet paraphrasing task. While the specific choice of pretraining dataset has minimal effect, the absence of pretraining results in a significant drop in quality.}
    \label{tab:tweet_pretraining}
    \vskip 0.15in
    \begin{small}
    \begin{adjustbox}{max width=\textwidth}
        \begin{tabular}{clcccc}
        \toprule
            \# embeddings & Metric & w/o pretraining & FineWeb & tweets & FineWeb + tweets \\
            \midrule
            \multirow{2}{*}{4} & Perplexity $\downarrow$ & 2.190 & 2.202 & \textbf{2.188} & 2.196 \\
                               & ROUGE-L $\uparrow$      & 0.435 & 0.432 & 0.434 & \textbf{0.436} \\
            \midrule
            \multirow{2}{*}{8} & Perplexity $\downarrow$ & 2.166 & 2.147 & \textbf{2.143} & 2.147 \\
                               & ROUGE-L $\uparrow$      & 0.430 & 0.444 & \textbf{0.447} & 0.445 \\

            \midrule
            \multirow{2}{*}{16} & Perplexity $\downarrow$ & 2.125 & 2.118 & 2.114 & \textbf{2.111} \\
                               & ROUGE-L $\uparrow$       & 0.451 & 0.450 & \textbf{0.455} & 0.451 \\
            \midrule
            \multirow{2}{*}{32} & Perplexity $\downarrow$ & 2.124 & 2.108 & 2.109 & \textbf{2.100} \\
                               & ROUGE-L $\uparrow$       & \textbf{0.454} & 0.443 & 0.447 & 0.447 \\
            \midrule
            \multirow{2}{*}{64} & Perplexity $\downarrow$ & 2.114 & \textbf{2.092} & 2.094 & 2.094 \\
                               & ROUGE-L $\uparrow$       & \textbf{0.452} & 0.449 & 0.450 & \textbf{0.452} \\
            \midrule
            \multirow{2}{*}{128} & Perplexity $\downarrow$ & 2.143 & 2.098 & 2.097 & \textbf{2.093} \\
                                 & ROUGE-L $\uparrow$      & 0.433 & 0.447 & \textbf{0.456} & 0.451 \\
            \midrule
            \multirow{2}{*}{Average} & Perplexity $\downarrow$ & 2.144 & 2.127 & 2.124 & \textbf{2.123} \\
                                     & ROUGE-L $\uparrow$      & 0.442 & 0.444 & \textbf{0.448} & 0.447 \\
        \bottomrule
        \end{tabular}
    \end{adjustbox}
    \end{small}
    \vskip -0.1in
\end{table*}

As demonstrated in prior work, pretraining can enhance compression performance \cite{ge2024incontext}. In this study, we aim to quantify its impact on ComMer and explore the effect of different pretraining datasets on the personalized tweet paraphrasing task. Specifically, in addition to pretraining on the FineWeb dataset, we experiment with pretraining on the training set of the personalized tweet paraphrasing task, where we apply the extended auto-encoding task to users' past tweets. We further investigate pretraining on FineWeb followed by pretraining on the personalized tweet paraphrasing dataset, as well as a no-pretraining baseline.
Table \ref{tab:tweet_pretraining} shows that pretraining improves downstream performance, measured by perplexity. Interestingly, the choice of pretraining dataset has minimal effect on the final results.

\subsection{Generalization of number of documents}

\begin{figure}
\vskip 0.2in
\begin{center}
\includegraphics[width=\columnwidth]{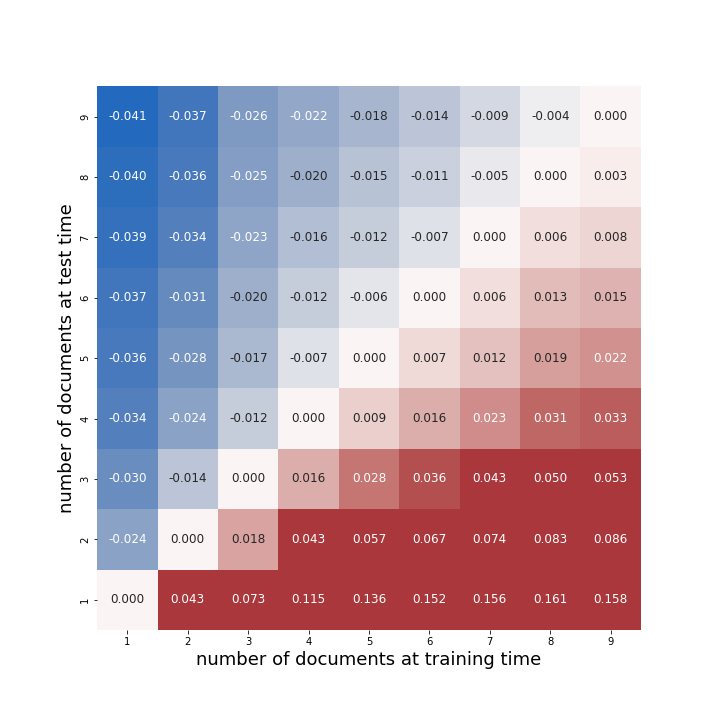}
\caption{Evaluation of ComMer on a different number of documents than those used during training, demonstrated using the personalized tweet paraphrasing task. Each cell shows the perplexity difference relative to the perplexity achieved when using the number of documents the model was trained for. ComMer's performance improves when exposed to more documents than it was trained on, but degrades when fewer documents are provided. The most intriguing case is when only a single document is used during training: Despite the compressor not being trained in tandem with the mean pool operation, averaging the compressions at test time still enhances performance, indicating that deep neural networks tend to operate linearly in the latent space.}
\label{fig:generalization_perplexity}
\end{center}
\vskip -0.2in
\end{figure}

In all previous experiments, we used the same number of documents at both training and test time. However, in realistic scenarios, users at test time may provide a different number of documents than what the model was trained on. While ComMer could be trained to handle a varying number of documents during training, it is valuable to explore how well ComMer generalizes to an out-of-distribution number of documents. Figure \ref{fig:generalization_perplexity} illustrates the perplexity\footnote{The ROUGE-L differences are too noisy to highlight any trend.} differences when the number of test documents differs from the training configuration, using 4 compression embeddings, demonstrated using the personalized tweet paraphrasing task. Specifically, let $perplexity_{ij}$ represent the perplexity of a model trained with $j$ documents when tested with $i$ documents. The $\langle i, j \rangle$ cell contains $perplexity_{ij} - perplexity_{jj}$. Interestingly, ComMer improves when exposed to more documents than it was trained on, while performance degrades when fewer documents are provided. The most intriguing case is $perplexity_{i1}$, where only a single document was used during training. In this scenario, the compressor was not trained in conjunction with the mean pool operation. Nevertheless, averaging the compressions at test time still enhances performance, suggesting that deep neural networks inherently operate linearly in the latent space, consistent with prior research findings \cite{elhage2022superposition,bricken2023monosemanticity,pmlr-v235-park24c}.

\subsection{Concatenation}

\begin{figure*}[t!]
\vskip 0.2in
\begin{center}
\includegraphics[width=1\linewidth]{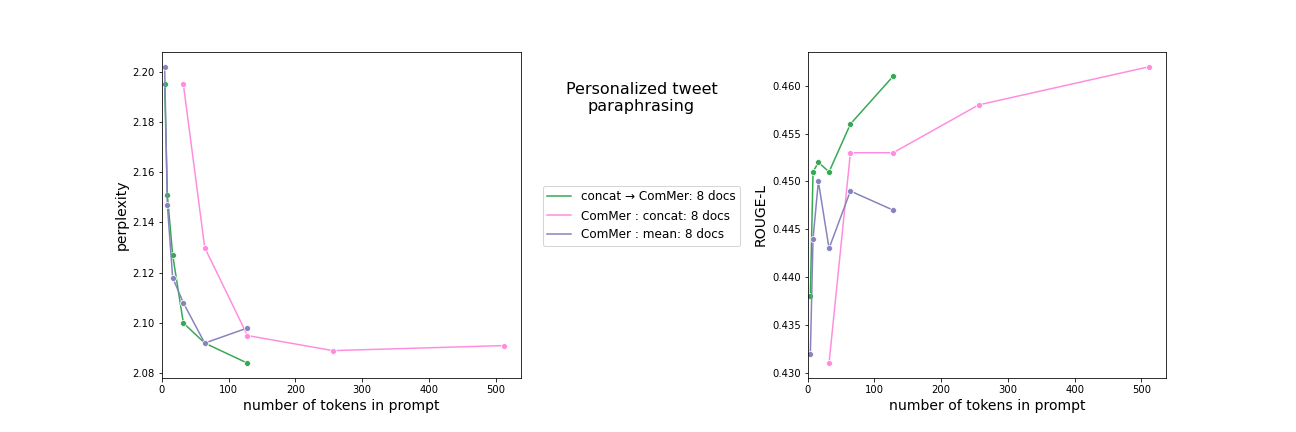}
\caption{Compariosn of different merging operations, demonstrated using the personalized tweet paraphrasing task. Concatenating compressions (\textit{ComMer : concat}) is less effective than averaging them (\textit{ComMer : mean}). Concatenating documents before compression (\textit{concat $\rightarrow$ ComMer}) yields marginal perplexity improvement, but increases compression computational costs.}
\label{fig:tweet_concatenation}
\end{center}
\vskip -0.2in
\end{figure*}

Our choice of mean pooling as the merging operation is motivated by the need to optimize two types of costs:
\begin{enumerate}
    \item \textbf{Token processing costs}: The costs associated with the number of tokens the frozen backbone model needs to process when generating the response.
    
    \item \textbf{Update costs}: The computational overhead involved in updating the representation of a set of documents when a new document is added.
\end{enumerate}
However, optimizing these costs can come on the expense of quality. To evaluate this trade-off, we explore two alternative approaches:
\begin{enumerate}
    \item \textbf{Concatenating the document compressions}: Instead of using mean pool, we concatenate the compressions. This approach increases the token processing costs, while keeping the update costs intact (compared to vanilla ComMer), assuming storage overhead is negligible.
    
    \item \textbf{Compressing the concatenation of documents}: This approach increases the update costs, while keeping the token processing costs intact.
\end{enumerate}
Figure \ref{fig:tweet_concatenation} compares these methods with vanilla ComMer, using the personalized tweet paraphrasing task. The results show that concatenating compressions (\textit{ComMer : concat}) is outperformed by vanilla ComMer (\textit{ComMer : mean}), indicating that the interference introduced by mean pooling is not detrimental to capturing a user's tweeting style.
Compressing the concatenation of documents (\textit{concat $\rightarrow$ ComMer}) yields marginal quality improvements over vanilla ComMer but incurs higher compression and update costs. Therefore, we argue that vanilla ComMer is the most cost-effective approach for document representation. Additionally, vanilla ComMer is better suited for retrieval-based strategies, where only relevant documents are used per query, a design that is not feasible with the document concatenation approach when compressions are pre-calculated. We leave further exploration of this approach to future work.

%% file: 06_discussion.tex
\section{Discussion}

In this work, we introduced ComMer, a novel and cost-effective approach for compressing multiple documents. We evaluated how our method compares to an existing popular approach - prompt-tuning - that exposes LLMs to document content directly. We demonstrated that compressing merging is beneficial in personalized skill learning tasks, but less effective in knowledge-intensive tasks.

The difference between the two task types intuitively makes sense: in the personalized skill learning tasks, a single document typically does not encapsulate all the information required to learn a user's style. Instead, each document suggests a range of possible styles. By intersecting these sets of potential styles across multiple documents, we can narrow down to those that best represent the user's overall style. When compressions are averaged, the particularities of each document are diminished, resulting in a representation that encapsulates the essential signal of the user’s style.

In knowledge-intensive tasks, the particularities of each document's compression are crucial, as they capture the specific details of individual documents. Averaging the compressions can negate these particularities through destructive interference, potentially explaining the degradation in quality when more documents are used. Although this averaging produces a single representation that broadly reflects the user's documents, it falls short when the goal is to preserve the detailed information contained in each document.

We conjecture that this qualitative difference between the two task types reflects a broader trend across personalized skill learning and knowledge-intensive tasks, and leave the exploration of additional tasks within these two categories for future research.

Furthermore, we hypothesize that improvements in single-document compression - whether through better hyperparameters, enhanced architectures or better weight optimization - will yield corresponding benefits in multi-document compression. In other words, our method and conclusions are not constrained to the specific compressor architecture implemented in this study, and future work may explore other compressor architectures.

%% file: 07_impact.tex
\section{Impact Statement}

LLMs have become remarkably capable in recent years. Personalizing them to users' needs and data is one of the next big challenges in making these models more accessible and helpful. ComMer facilitates the efficient personalization of LLMs to user data, which can have several positive societal consequences:

\begin{itemize}
\item \textbf{Increased accessibility:} It enables wider access to personalized LLMs, particularly for users with limited data, for whom training a personalized model may yield low quality due to a small training set. 
\item \textbf{Reduced environmental impact:} The higher efficiency of ComMer translates to lower energy consumption, contributing to a reduced carbon footprint. 
\end{itemize}

However, it's equally important to acknowledge the potential ethical concerns and negative societal consequences:

\begin{itemize}
\item \textbf{Bias:} If the user data used to personalize the LLM contains biases, ComMer might preserve these biases, leading to discriminatory or unfair outputs. Future work should aim to compress data in a way that captures only relevant and non-discriminatory signals. 
\item \textbf{Privacy:} Training ComMer on real user data raises privacy concerns, especially if the data contains sensitive personal information. It is essential to ensure robust anonymization and security measures are in place or that high-quality synthetic data is used. 
\item \textbf{Data compliance:} Models personalized to user data should comply with data privacy and data security laws and standards. 
\end{itemize}

In this paper we do not aim to address these challenges. However, addressing them is crucial for the responsible development and deployment of techniques like ComMer.

%% file: 08_appendix.tex
\section{Appendix}

\subsection{Implementation details}
\label{section:implementation_details}
\textbf{Data}
For the personalized tweet paraphrasing and the personalized news headline generation tasks, we employ BM25 to retrieve relevant documents, following the methodology used by LaMP. To ensure computational efficiency, we exclude examples exceeding 2900 characters, including those in the PerLTQA task.

\textbf{Model}
We utilize the Hugging Face implementation of \texttt{gemma-2b-it} and the \texttt{peft} package for integrating LoRA. The LoRA configuration includes a rank of 64, alpha of 16, and LoRA-dropout of 0.1.

\textbf{Optimization}
For training we employ a cosine learning rate scheduler with the first 3\% of steps allocated for warm-up. The maximum learning rate for prompt-tuning is $3 \times 10^{-2}$. For ComMer, the maximum learning rates is $1 \times 10^{-4}$ for the LoRA weights and $1 \times 10^{-2}$ for the compression embeddings. We observed that using a homogeneous learning rate is sub-optimal. We use AdamW optimizer with a weight decay of 0.001 and a maximum gradient norm (for gradient clipping) of 0.3. Batch sizes are set to 16 for LaMP tasks and 1 for PerLTQA. Training proceeds for 20 epochs for LaMP tasks and 1 epoch of 80K examples for PerLTQA. We early stop according to perplexity based on a small reserved validation set of 1024 examples.

\textbf{Evaluation}
For ease of research, we use LaMP's official development set as the test set due to the unavailability of test set labels. For evaluation, we use greedy decoding to compute ROUGE metrics.

\subsection{Dataset examples}
\label{section:dataset_examples}
Following are examples from each of the datasets used in this study, when setting the number of documents to two.

\begin{table}[b]
    \centering
    \caption{An example from the personalized tweet paraphrasing task.}
    \vskip 0.15in
    \begin{small}
    \begin{adjustbox}{max width=\textwidth}
        \begin{tabular}{p{0.8\linewidth}}
        \toprule
            \textbf{Document 1:} \\
            i need a new myspace song. any suggestions ? \\
            \midrule
            \textbf{Document 2:} \\
            phones gonna die. I'll try and get on the comp soonish.. \\
            \midrule
            \textbf{Question:} \\
            Paraphrase the following tweet without any explanation before or after it: I require new photos and music for my phone. \\
            \midrule
            \textbf{Answer:} \\
            still need new pics on my phone\ \ and music.. \\
        \bottomrule
        \end{tabular}
    \end{adjustbox}
    \end{small}
    \vskip -0.1in
\end{table}

\begin{table}[b]
    \centering
    \caption{An example from the personalized news headline generation task.}
    \vskip 0.15in
    \begin{small}
    \begin{adjustbox}{max width=\textwidth}
        \begin{tabular}{p{0.8\linewidth}}
        \toprule
            \textbf{Document 1:} \\
            Text: Beat the summer heatwave with this ridiculously simple trick that keeps your vodka ice-cold for hours.

Title: The Coolest Thing You'll Do With Vodka \\
            \midrule
            \textbf{Document 2:} \\
            Text: A few quick tips to selecting fresh fish and why buying local will taste a whole lot better.

Title: How Fresh Is Your Fish? \\
            \midrule
            \textbf{Question:} \\
            Generate a headline for the following article: A simple and straightforward workaround for the messiness of roasting beets. \\
            \midrule
            \textbf{Answer:} \\
            The Clean \& Easy Way to Roast Beets \\
        \bottomrule
        \end{tabular}
    \end{adjustbox}
    \end{small}
    \vskip -0.1in
\end{table}

\begin{table}[b]
    \centering
    \caption{An example from the PerLTQA task.}
    \vskip 0.15in
    \begin{small}
    \begin{adjustbox}{max width=\textwidth}
        \begin{tabular}{p{0.8\linewidth}}
        \toprule
            \textbf{Document 1:} \\
            Long Meili and Zhang Yang are a romantic couple who support and encourage each other and pursue their dreams together. \\
            \midrule
            \textbf{Document 2:} \\
            Long Meili and Wang Li collaborated on a new topic in the field of sports medicine. \\
            \midrule
            \textbf{Question:} \\
            What is the interaction between Long Meili and Zhang Yang? \\
            \midrule
            \textbf{Answer:} \\
            They support, encourage and pursue their dreams together. \\
        \bottomrule
        \end{tabular}
    \end{adjustbox}
    \end{small}
    \vskip -0.1in
\end{table}

\begin{table}[b]
    \centering
    \caption{An example from the FineWeb task.}
    \vskip 0.15in
    \begin{small}
    \begin{adjustbox}{max width=\textwidth}
        \begin{tabular}{p{0.8\linewidth}}
        \toprule
            \textbf{Document 1:} \\
            Hi! Within the last month when I add an event to my google calendar it will change the title to "Fake event to work around a calendar issue" and if I try to edit it, it will disappear within a minute or two. Anyone have any ideas? Thanks!! \\
            \midrule
            \textbf{Document 2:} \\
            Hey! My names Sydnie and i guess im 16 now? you can call me syd or even cici is fine. I draw, cosplay, and play games. Oh and homework, i do that too. \\
            \midrule
            \textbf{Question:} \\
            What does document 2 contain? \\
            \midrule
            \textbf{Answer:} \\
            Hey! My names Sydnie and i guess im 16 now? you can call me syd or even cici is fine. I draw, cosplay, and play games. Oh and homework, i do that too. \\
        \bottomrule
        \end{tabular}
    \end{adjustbox}
    \end{small}
    \vskip -0.1in
\end{table}